\documentclass[10pt,twocolumn,letterpaper]{article}

\usepackage[final]{cvpr}      

\usepackage{graphicx}
\usepackage{amsmath}
\usepackage{amssymb}
\usepackage{booktabs}

\usepackage[pagebackref,breaklinks,colorlinks]{hyperref}

\usepackage[capitalize]{cleveref}
\crefname{section}{Sec.}{Secs.}
\Crefname{section}{Section}{Sections}
\Crefname{table}{Table}{Tables}
\crefname{table}{Tab.}{Tabs.}

\def\cvprPaperID{7} 
\def\confName{CVPR}
\def\confYear{2023}

\begin{document}

\title{Face Transformer: Towards High Fidelity and Accurate Face Swapping}

\author {
    Kaiwen Cui, 
    Rongliang Wu,
    Fangneng Zhan,
    Shijian Lu\thanks{corresponding author.} \\
    Nanyang Technological University, Singapore
    \\
    {\tt\small \{Kaiwen001, Rongliang001\}@e.ntu.edu.sg, 
    \{fnzhan,Shijian.Lu\}@ntu.edu.sg} \\
}

\twocolumn[{
\renewcommand\twocolumn[1][]{#1}
\maketitle
\begin{center}
    \centering
    \includegraphics[width=\linewidth]{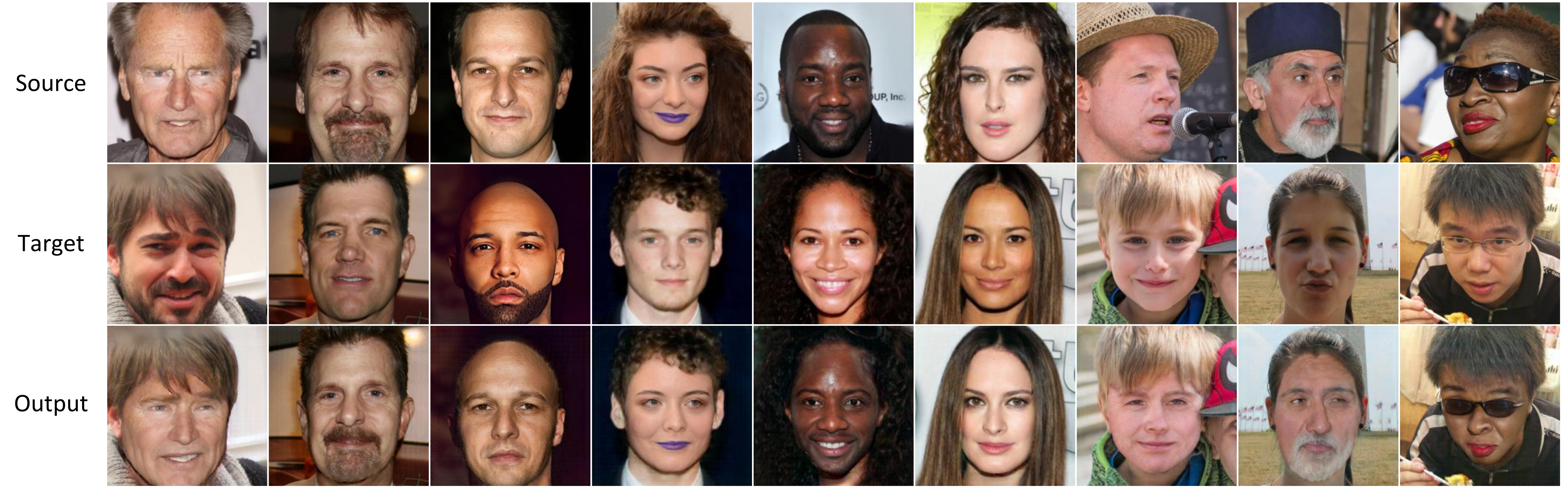}
    \vspace{-15pt}
    \captionof{figure}{The proposed Face Transformer fuses the identity of $Source$ face and attributes of $Target$ faces according to the semantic-aware correspondence that is constructed by using a transformer architecture. It produces high-fidelity and accurate face swapping over multiple public datasets including CelebA-HQ \cite{liu2018large}, VGGFace \cite{parkhi2015deep} and FFHQ \cite{karras2019style} (Sample images in columns 1-3, 4-6, and 7-9 are from CelebA-HQ, VGGFace, and FFHQ, respectively).}
    \label{fig1}
\end{center}
}]

\renewcommand{\thefootnote}{\fnsymbol{footnote}}


\begin{abstract}
Face swapping aims to generate swapped images that fuse the identity of source faces and the attributes of target faces. Most existing works address this challenging task through 3D modelling or generation using generative adversarial networks (GANs), but 3D modelling suffers from limited reconstruction accuracy and GANs often struggle in preserving subtle yet important identity details of source faces (e.g., skin colors, face features) and structural attributes of target faces (e.g., face shapes, facial expressions). This paper presents Face Transformer, a novel face swapping network that can accurately preserve source identities and target attributes simultaneously in the swapped face images. We introduce a transformer network for the face swapping task, which learns high-quality semantic-aware correspondence between source and target faces and maps identity features of source faces to the corresponding region in target faces. The high-quality semantic-aware correspondence enables smooth and accurate transfer of source identity information with minimal modification of target shapes and expressions. In addition, our Face Transformer incorporates a multi-scale transformation mechanism for preserving the rich fine facial details. Extensive experiments show that our Face Transformer achieves superior face swapping performance qualitatively and quantitatively.
\end{abstract}

\section{Introduction}
\label{sec:intro}
Face Swapping aims to generate new face images that combine the source faces' identities which include skin colors, face features, makeups and etc., as well as the target faces' attributes that include head poses, head shapes, facial expressions, backgrounds, etc. Automated and realistic face swapping has attracted increasing interest in recent years thanks to its wide range of applications in different areas, such as movie composition, computer games and privacy protection, etc. However, it remains a challenging task to accurately and realistically extract and fuse identity information from source faces and attribute features from target faces.

Most existing face swapping methods can be broadly classified into two categories which exploit 3D face models and generative adversarial networks (GANs)\cite{goodfellow2014generative}, respectively. 
Earlier works \cite{lin2012face, nirkin2018face} employ 3D models to deal with the pose and perspective differences between source and target faces, which estimate 3D shapes of source and target faces and use the estimated 3D shapes as proxy for face swapping. However, 3D based methods suffer from limited accuracy in 3D reconstruction which tend to generate various distortions and artefacts in the swapped face images. Inspired by the great success of GANs \cite{goodfellow2014generative, dcgan, CGAN,cyclegan}, several works \cite{bao2017cvae, korshunova2017fast, natsume2018fsnet, natsume2018rsgan, nirkin2019fsgan, li2019faceshifter} employ generative models and have achieved very impressive face swapping performance. Though GAN-based networks can generate high-fidelity swapping, they still struggle in preserving subtle yet important identity details of source faces (in skin colors, face features, makeups, etc.) and structural attribute features of target faces (in face shapes, facial expressions, etc.).
 
This paper presents Face-Transformer, an innovative face swapping network that achieves superior face swapping by accurate preservation of identity details of source faces and structural attributes of target faces. The superior swapping performance is largely attributed to a transformer architecture we introduce into the face swapping task. Specifically, we employ the transformer to build up semantic-aware correspondence between source and target faces with which the identity features of source faces can be mapped to the corresponding region of target faces smoothly and accurately. The construction of semantic-aware correspondence can not only preserve structural attributes of target faces but also achieve high-fidelity identity of source faces especially for subtle yet important details as illustrated in Fig. \ref{fig1}. The proposed Face Transformer consists of three modules including a face parsing module, a face feature transformation module (FFTM) and a face generation module (FGM). Face parsing module is an off-the-shelf module\footnote{\url{https://github.com/zllrunning/face-parsing.PyTorch}} which predicts face masks to separate inner faces from the background and extracts face semantics to guide the training of transformer in FFTM. FFTM takes source face, target face and target face semantics as inputs to learn semantic-aware correspondence between two faces, and maps the identity features of source faces to the corresponding regions of target faces. FGM finally generates high-fidelity swapped face images based on the transformation of multi-scale facial features by FFTM.

The contributions of this work are threefold.
\textit{First}, we design Face-Transformer, an innovative network that achieves accurate face swapping by introducing transformer into face swapping task. The transformer learns semantic-aware correspondence between source and target faces which facilitates the feature transfer from source faces to target faces smoothly. To the best of our knowledge, this is the first work that introduces transformer architectures for the face swapping task.
\textit{Second}, we propose a novel multi-scale feature transformation strategy that helps learn more powerful feature representation and achieve more accurate face swapping.
\textit{Third}, extensive experiments show that the proposed Face Transformer achieves superior face swapping quantitatively and qualitatively.

\begin{figure*}[t]
\centering
\includegraphics[width=0.9\linewidth]{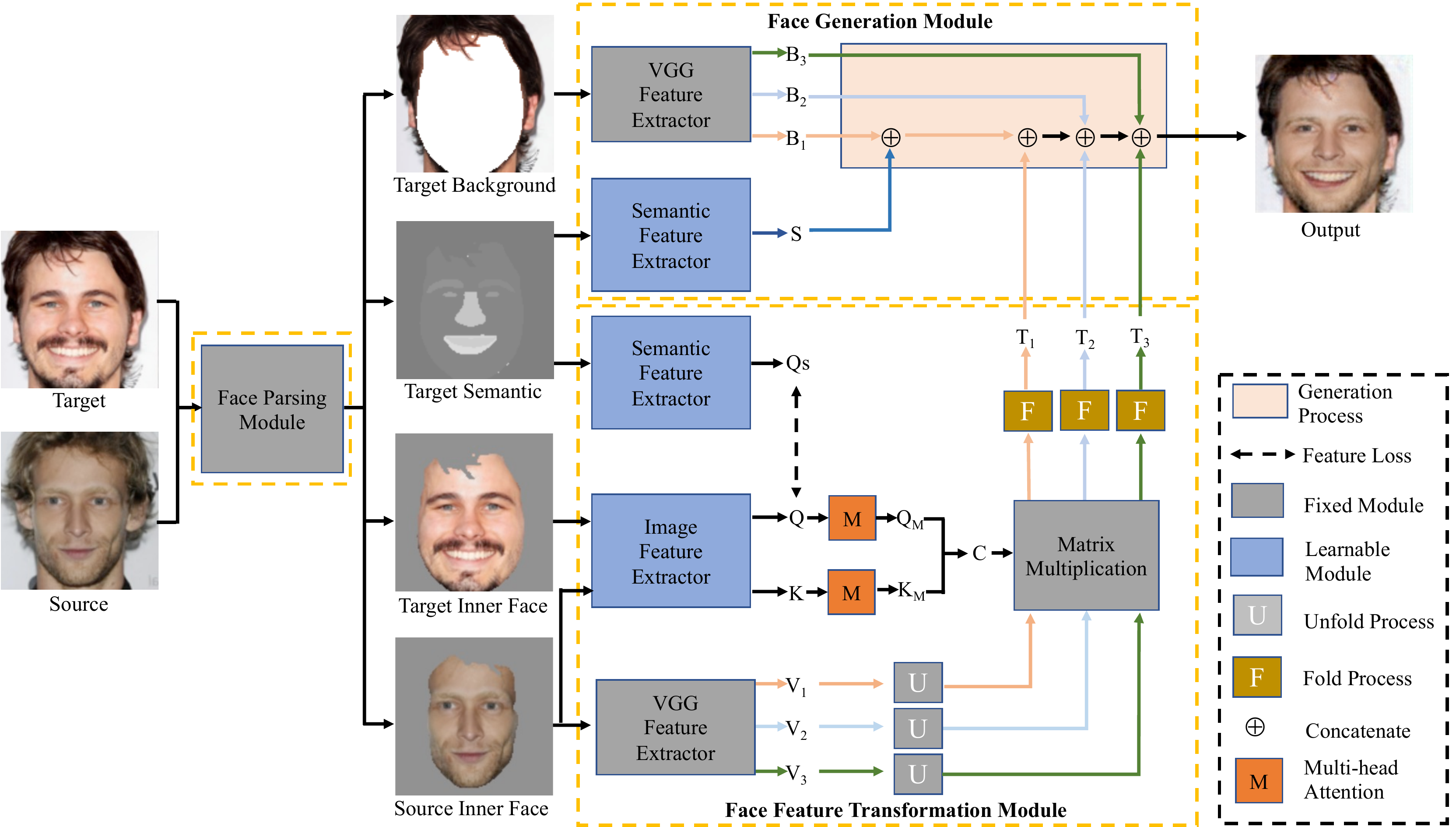}
\caption{ 
The architecture of the proposed Face Transformer: Given a \textit{Source} face and a \textit{Target} face, a \textit{Face Parsing Module} first extracts a $Source$ $Inner$ $Face$, a $Target$ $Inner$ $Face$, a $Target$ $Semantic$ map, and a $Target$ $Background$, respectively. A transformer then learns a semantic-aware correspondence matrix $C$ from source and target face features $K$ and $Q$. The learned $C$ maps multi-scale identity features $V_1$, $V_2$, and $V_3$ of the source face (extracted by a pre-trained $VGG$ $Feature$ $Extractor$) to $T_1$, $T_2$, and $T_3$ that are aligned with the corresponding region of the target face. Finally, $T_1$, $T_2$, and $T_3$ are concatenated with multi-scale background features $B_1$, $B_2$, and $B_3$ of the target face to generate the swapped face $Output$. 
}
\label{im_stru}
\end{figure*}

\section{Related Works}

\subsection{Face Swapping} 

Face swapping has achieved remarkable progress in recent years. Existing face swapping methods can be broadly classified in two categories: 3D-based methods and GAN-based methods.

\subsubsection{3D-Based methods}
Early face swapping task involves manual intervention \cite{blanz2004exchanging} until the introduction of automated methods \cite{bitouk2008face}. However, automated methods cannot preserve face expressions well and Face2Face \cite{thies2016face2face} was then proposed to transfer expressions from target faces to source faces. Face2Face works by applying 3D morphable face model (3DMM) to source faces and target faces and then transferring the expression component from target face to source face. To further preserve the occlusions, Nirkin \etal \cite{nirkin2018face} collected data to train an occlusion-aware face segmentation network in a supervised way. However, the 3D-Based models suffer from limited accuracy in 3D reconstruction which tend to generate various distortions and artefacts in the swapped face image.

\subsubsection{GAN-Based methods}

Generative adversarial network~\cite{goodfellow2014generative} (GAN) has achieved great success in image generation~\cite{koksal2020rf,zhan2021unbalanced, cui2022genco,yu2022towards,huang2022masked,zhan2022bi,zhan2022marginal,zhan2019spatial}. 
Leveraging the fast development of GANs, several methods \cite{ bao2017cvae, korshunova2017fast, natsume2018fsnet, natsume2018rsgan, nirkin2019fsgan, li2019faceshifter} introduce GANs for face swapping and have achieved quite impressive progress.

Deepfakes\footnote{\url{https://github.com/ondyari/FaceForensics/tree/master/dataset/DeepFakes}} and Faceswap \footnote{\url{https://github.com/ondyari/FaceForensics/tree/master/dataset/FaceSwapKowalski}}have achieved great success in face swapping recently, but they need to train a new model with two video sequences for each new input. To mitigate this constraint, subject-agnostic face swapping has attracted increasing interest. For example, Natsume \etal \cite{natsume2018rsgan} disentangles the embedding of face and hairs and recombines them to generate swapped face. Natsume \etal \cite{natsume2018fsnet} utilizes a latent space to preserve face identity in source face and appearance of hair style and background region in target face. Nirkin \etal \cite{nirkin2019fsgan} utilizes an occlusion-aware face segmentation network to preserve facial occlusion and Li \etal \cite{li2019faceshifter} presents a heuristic error acknowledgement refine network. Although GAN-based methods can achieve high fidelity face swapping, most of them struggle in preserving subtle yet important identity details of source faces (in skin color, face features, makeups, etc.) and structural attributes of target face (in face shape, face expression, etc.). The proposed Face Transformer learns a semantic-aware correspondence between source faces and target faces that can accurately and smoothly transfer source identity information with minimal modification of the shapes and expressions of target faces.

\subsection{Transformer}
Initially proposed for natural language processing tasks, transformer has recently become an emerging component that is widely applied in various vision tasks. With its superiority over convolutional neural networks (CNNs) in capturing long-distance relationship, transformer-based vision frameworks have demonstrated their effectiveness in image classification\; \cite{dosovitskiy2020image,liu2021swin,deit}, object detection\; \cite{DETR,DeformableDETR,MetaDETR}, image synthesis \cite{esser2020taming,chen2020pre,zhan2022auto,yu2021diverse}, super-resolution\cite{yang2020learning}, etc. The key in the transformer architecture is the attention mechanism \cite{vaswani2017attention} that models interactions between its inputs regardless of their relative position to one another.

In this work, we exploit the strong expressivity of the transformer architecture for the challenging face swapping task, where the major challenge is to construct accurate semantic-aware correspondence between source and target faces. The superior performance achieved by the proposed Face Transformer demonstrate that the transformer architecture is a natural fit for this task.

\section{Methods}

As illustrated in Fig. \ref{im_stru}, the proposed Face Transformer consists of three modules including a face parsing module, a face feature transformation module (FFTM) and a face generation module (FGM). The face parsing module is an off-the-shelf face parsing model that produces face masks and face semantics, which separate inner faces from the image background and guide the learning of semantic-aware correspondence to be used in FFTM. Once the inner face and face semantics are obtained,
FFTM learns the semantic-aware correspondence between source inner face and target inner face and transforms multi-scale features of source inner face to corresponding regions of target faces based on the learnt semantic-aware correspondence. The multi-scale transformed features are then progressively concatenated with features of target semantic and multi-scale features of the target-face background for the generation swapped face images. FFTM and FGM are trained in an end-to-end manner as to be discussed in the ensuing subsections.

\subsection{Face Feature Transformation Module}  
\label{face feature transformation module}

With the extracted source inner faces, target inner faces and target semantics, FFTM learns to map identity features of the source inner faces to target inner faces according to the semantic-aware correspondence between the source and target face features.

\subsubsection{Feature extractor}
FFTM has three feature extractors. The first is a VGG feature extractor which employs a pre-trained VGG-19 model to extract multi-scale features $V_1\in\mathcal{R}^{H*W*C}$, $V_2\in\mathcal{R}^{2H*2W*\frac{C}{2}}$, and $V_3\in\mathcal{R}^{4H*4W*\frac{C}{4}}$ from the source inner face. The extracted features are then mapped to the corresponding regions of target face based on the learnt semantic-aware correspondence. The second extractor is a learnable image feature extractor which extracts features of source inner face (K) and target inner face (Q). The third is a learnable semantic feature extractor which extracts features from the target semantic map ($Q_S$) for guiding the generation of semantic-aware target inner face feature Q. The incorporation of the learnable semantic feature extractor is based on the observation that target semantics capture facial expression and shape information which is critical in face swapping. The two learnable feature extractors take similar network architecture as in \cite{zhang2020cross}.

\subsubsection{Feature Transformation}
The major component of FFTM is a transformer that first performs multi-head attention over source face features $K$ and target face features $Q$. This produces the corresponding feature representation $Q_M$ and $K_M$. The transformer then performs cross attention between $Q_M$ and $K_M$ which builds up connections between the input $K$ and output $Q$. 

Similar to \cite{vaswani2017attention}, the multi-head attention over $Q$ and $K$ is formulated as $Q_M = [head_1,...,head_h]W_0$
and 
\begin{align}\begin{aligned}
head_i =  softmax(\frac{(QW_i^Q)(KW_i^K)^T}{|QW_i^Q||(KW_i^K)|}),
\end{aligned}\end{align}
where $W$ denotes learnable parameters. $K_M$ is obtained similarly. 

Different from the cross attention in typical transformers, we first employ $Q_M$ and $K_M$ to learn a correspondence matrix $C$ that represents the semantic-aware correspondence between source face features and target face features. The learned semantic-aware correspondence matrix is then applied over the facial features that are extracted from the VGG feature extractor. Specifically, the semantic-aware correspondence matrix $\mathcal{C}$ can be obtained by performing dot products of each channel-wise feature in $Q_M$ with all channel-wise feature in $K_M$ and applying softmax over the dot-product results:
\begin{align}\begin{aligned}
\mathcal{C} = softmax{(\frac{Q_MK_M^T}{|Q_M||K_M|})}.
\end{aligned}\end{align}

\subsubsection{Multi-scale feature transformation}
We also design a multi-scale feature transformation by transforming $V_1\in\mathcal{R}^{H*W*C}, V_2\in\mathcal{R}^{2H*2W*\frac{C}{2}}, V_3\in\mathcal{R}^{4H*4W*\frac{C}{4}}$ simultaneously by using the learned correspondence matrix. To resolve the spatial discrepancy across multi-scale features, we apply an unfold process $U$ to each feature which unifies the spatial dimension of all features to be H*W. The unfolded features are then multiplied with the correspondence matrix to produce transformed features. Finally, we restore the spatial dimension of each feature by a fold process $F$. We design the multi-scale feature transformation with two major purposes. First, transforming multi-scale features helps preserve more fine facial details of source faces in the swapped face image. Second, learning the multi-scale feature transformation can inversely help improve the learning of the semantic-aware correspondence matrix.

\begin{figure}[t]
\centering
\includegraphics[width=1\linewidth]{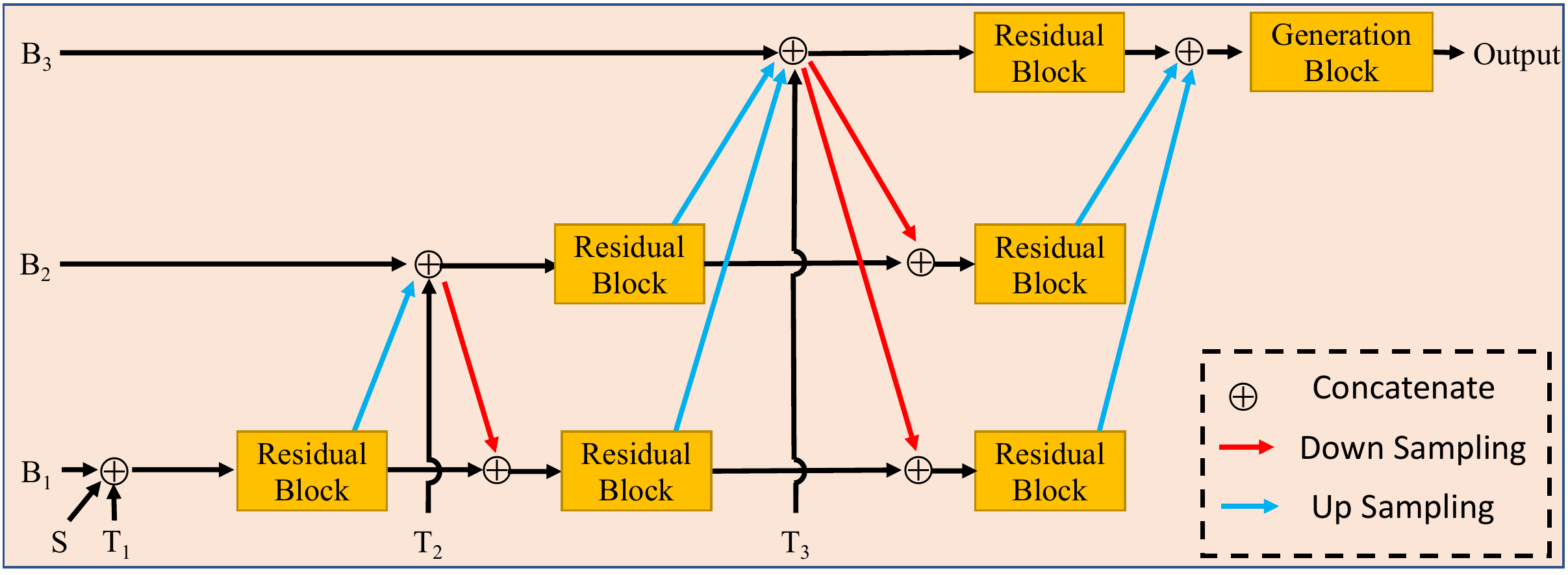}
\caption{
The face generation details: Taking semantic feature $S$, transformed multi-scale face features $T_1$, $T_2$, $T_3$ and multi-scale background features $B_1$, $B_2$, and $B_3$ as input, the face generator progressively and pair-wisely exchanges information among features of different scales. It handles feature scale differences by up-sampling (with an interpolation layer, a convolution layer and a ReLU layer) and down-sampling (with a convolution layer and a ReLU layer). Residual block has the same architecture as \cite{park2019semantic} and the generation block contains a convolution layer and a tanh layer.
}
\label{generator}
\end{figure}
\subsection{Face Generation Module}

Face generation module is used to fuse the transformed multi-scale features from face feature transformation module (FFTM) as well as multi-scale features of the target backgrounds ($B_1, B_2$, and $B_3$) and synthesize final high fidelity face images. To preserve the expressions and shapes of target faces, we employ another learnable semantic feature extractor to extract semantic features as input as well. 
The detailed architecture of the face generation module is shown in Fig. \ref{generator}. 
Inspired by \cite{yang2020learning} that exchanging information between features of different scales in generation can assist generator in learning more powerful feature representations and preserve better texture details, our face generation module progressively and pair-wisely exchanges information between features of different scales as illustrated in Fig. \ref{generator}.

\begin{table}
\begin{center}
\resizebox{0.9\linewidth}{!}{
\begin{tabular}{c c c c c c}
\hline
Method  & ID $\downarrow$ & Pose $\downarrow$ &  Expression $\downarrow$ & Shape $\downarrow$ &  Quality $\uparrow$ \\
\hline\hline
DeepFake & 38.26 & 4.14  & 43.04 & 66.3 & 0.95\\
FaceSwap & 37.29  & \textbf{2.51} & 29.1 & 47.8  & 0.972\\
FaceShifter & 38.73 & 2.96 &  27.93 & 51.4 & 0.977\\
Ours & \textbf{34.49} & 3.13 & \textbf{23.74}  & \textbf{44.3} &  \textbf{0.978}\\

\hline
\end{tabular}}
\end{center}
\vspace{-3mm}
\caption{
Quantitative comparisons of the proposed Face Transformer ($Ours$) with state-of-the-art methods $DeepFakes $, $FaceSwap$ and $FaceShifter$ \cite{li2019faceshifter} over the public dataset Faceforensics++ \cite{rossler2019faceforensics++}}
\label{quantitative}
\end{table}

\begin{figure*}[t]
\centering
\includegraphics[width=0.8\linewidth]{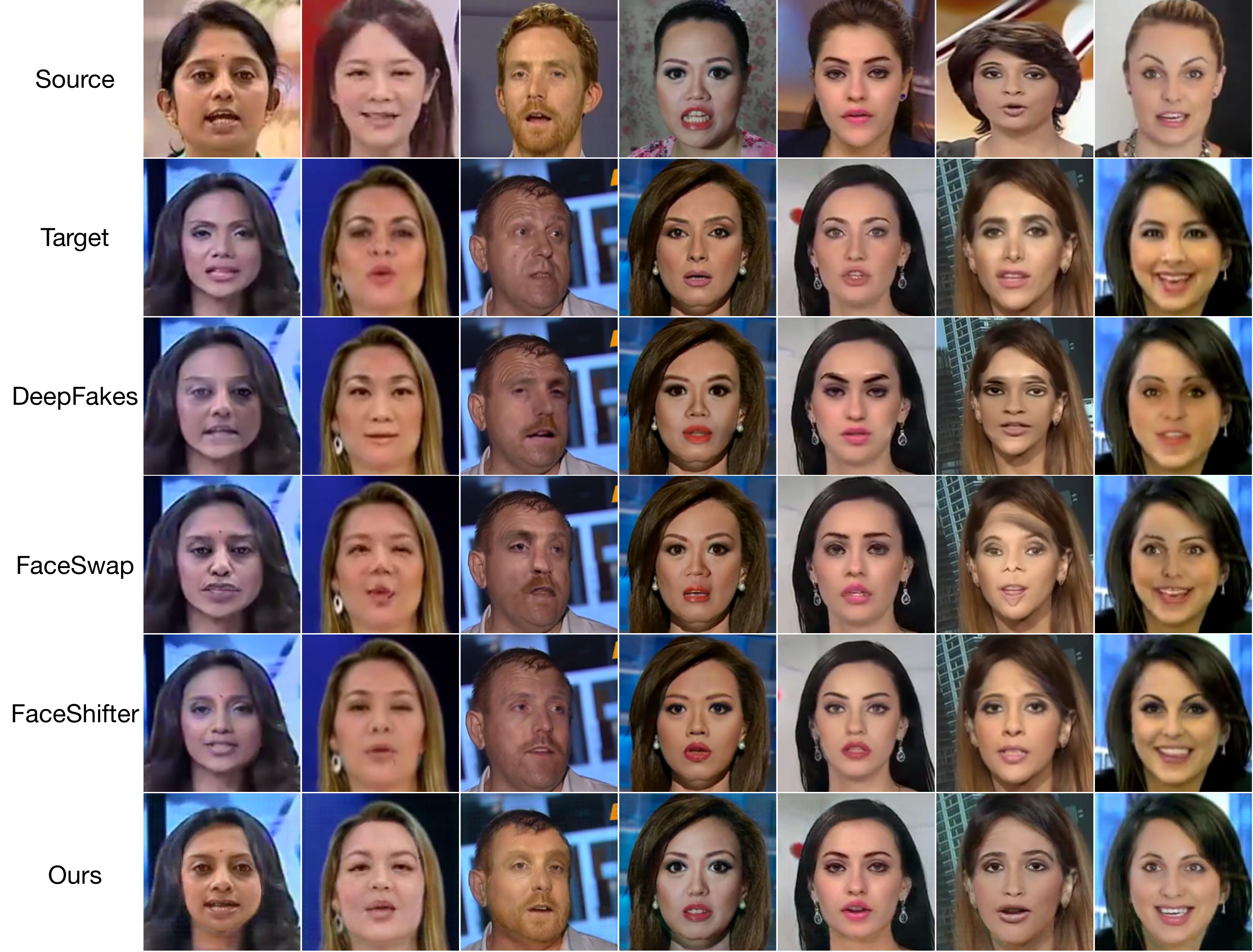}
\caption{
Qualitative comparisons of the proposed Face Transformer ($Ours$) with state-of-the-art methods $DeepFakes $, $FaceSwap$ and $FaceShifter$ \cite{li2019faceshifter} over the public dataset Faceforensics++ \cite{rossler2019faceforensics++}.
}
\label{qualitative}
\end{figure*}

\subsection{Training Objectives}

As mentioned in Section \ref{face feature transformation module}, we employ L1 loss between Q and $Q_S$ to preserve semantic information in Q. The feature loss $\mathcal{L}_{f}$ between Q and $Q_S$ is formulated by:
\begin{align}\begin{aligned}
\mathcal{L}_{f}= ||Q - Q_S||_1.
\end{aligned}\end{align}

\noindent To synthesize visually realistic images, the finally swapped face images ($f_{swap}$) are generated under adversarial loss $\mathcal{L}_{adv}$: 
\begin{align}\left.\begin{aligned}
\mathcal{L}_{adv}=\mathop{min}\limits_{G}\mathop{max}\limits_{D}\mathbb{E}[logD(f_{real})] + \mathbb{E}[log(1-D(f_{swap})], 
\end{aligned}\right.\end{align}
where $f_{real}$ is real face image. 

Face swapping task requires the swapped face ($f_{swap}$) to have the same shapes and expressions as the target face $f_{tgt}$. $f_{swap}$ and $f_{tgt}$ should therefore be semantic alike. Inspired by \cite{johnson2016perceptual}, we adopt the perceptual loss $\mathcal{L}_{perc}$ to minimize the semantic discrepancy: 
\begin{align}\left.\begin{aligned}
\mathcal{L}_{perc} = ||\phi_{l}(f_{swap}) - \phi_{l}(f_{tgt})||_1.
\end{aligned}\right.\end{align}

\noindent Following \cite{zhang2020cross}, we choose $\phi_{l}$ to be the activation after relu4\_2 layer in the VGG-19 network as this layer is highly semantic-related. 

In addition, we adopt the contextual loss $\mathcal{L}_{context}$  \cite{mechrez2018contextual} to align the face identities (in face color, texture, etc.) of the swapped face $f_{swap}$ with the source face $f_{src}$:
\begin{align}\left.\begin{aligned}
\mathcal{L}_{context} = \sum_{l}-log(CX(\phi_{l}(f_{swap}*m_{tgt}), \phi_{l}(f_{src}*m_{src}))),
\end{aligned}\right.\end{align}
where $m_{tgt}$ and $m_{src}$ are the mask of target face and source face that are produced by the face parsing network. Since the background of the swapped face shall be the same as the background of the target face, we employ respective masks to the swapped faces and the source faces to filter out background information and apply the mask of the target face to the swapped face (as they are ideally the same). The details of contextual similarity CX is defined in \cite{mechrez2018contextual}. We choose $\phi_{l}$ to be the activation after relu3\_2,  relu4\_2 and relu5\_2 as low-level features are more style-related.

\smallskip
\noindent The overall loss function of the proposed Face Transformer is 
\begin{align}\begin{aligned}
  \mathcal{L} = \lambda_{1}\mathcal{L}_{f} + \lambda_{2}\mathcal{L}_{adv} + \lambda_{3}\mathcal{L}_{perc} + \lambda_{4}\mathcal{L}_{context}. 
\end{aligned}\end{align}

We follow settings in \cite{zhang2020cross} to set $\lambda_{2}$ = 10, $\lambda_{3}$=0.001 and $\lambda_{4}$ = 1. We then empirically adjust  $\lambda_{1}$ to be 5.

\noindent

\section{Experiments}

\subsection{Datasets and Settings}

Similar to existing works\cite{li2019faceshifter, nirkin2019fsgan}, we evaluate and compare the proposed Face Transformer with state-of-the-art face swapping methods DeepFakes, FaceSwap and FaceShifter \cite{li2019faceshifter} over the public dataset Faceforensics++ \cite{rossler2019faceforensics++}. DeepFakes and FaceSwap are trained directly with Faceforensics++, while FaceShifter and the proposed Face Transformer are subject agnostic. Specifically , Faceshifter is trained by using the three datasets CelebA-HQ~\cite{liu2018large}, FFHQ~\cite{karras2019style} and VGGFace~\cite{parkhi2015deep}
while the proposed Face Transformer is trained over CelebA-HQ \cite{liu2018large} only. 
CelebA-HQ is a large-scale face image dataset that has 30K high-resolution face images. FFHQ consists of 70K high-quality images that contains considerable variations interms of age, ethnicity and image background. VGGFace consists of 2.6M high-resolution images that distribute over 2.6k identities. 

Note Faceforensics++ \cite{rossler2019faceforensics++} is a public dataset which contains 1000 videos of different identities. In our experiments, we follow \cite{li2019faceshifter} to evenly sample 10 frames from each video to form a test set, which consists of 10K face images.

\subsection{Implementation Details}

Following \cite{nirkin2019fsgan, li2019faceshifter}, we use five-point landmarks \cite{chen2014joint} to crop and align face images. The cropped images are resized to 256 $\times$ 256 in model training.
We use PyTorch \cite{paszke2017automatic} to implement the proposed Face Transformer. Adam optimizer \cite{kingma2014adam} is adopted as the optimizer with $\beta_{1}$=0.5 and $\beta_{2}$= 0.999.
The batch size is set to 4. We train the model for 20 epochs with a fixed learning rate of 0.0002.

\subsection{Evaluation Metrics}

We perform quantitative evaluations and comparisons with several metrics that have been used in prior research. The metrics include \textit{ID verification}, \textit{pose}, \textit{expression}, \textit{shape} and \textit{quality}.

\subsubsection{ID verification}
ID verification metric examines whether the swapped face images preserve the identity information of the source faces. It is evaluated according to the Euclidean distance between the identity features of the swapped and source face images that are extracted by a pre-trained face recognition model \cite{wang2018cosface}. 
Smaller distances indicate better identity preservation. We did not follow \cite{nirkin2019fsgan} that uses dlib to extract identity feature  as \cite{wang2018cosface} has much better performance in face recognition.

\subsubsection{Poses}
Pose metric evaluates how the swapped faces preserve the head pose of the target faces. It is computed by the Euclidean distance between head pose of target faces and the swapped faces. Following \cite{li2019faceshifter}, the head pose of each face image is estimated by \cite{ruiz2018fine}. Smaller distances indicate better pose preservation.

\subsubsection{Expressions}
Expression metric examines how the swapped faces preserve the expressions of the target faces. It is evaluated by the Euclidean distance between 2D landmarks of the target faces and the swapped faces as in \cite{nirkin2019fsgan}. In our experiments, we use open-source software dlib \cite{king2009dlib} to detect facial landmarks. Smaller distances indicate better expression preservation.

\subsubsection{Shape}
Shape metric examines how the swapped faces preserve the facial shape of target faces. As state-of-the-art methods do not compare this metrics, we design it ourselves by computing the Euclidean distance between target faces' mask and swapped faces' mask. We extract face masks by using the segmentation network in \cite{nirkin2019fsgan}. Smaller distances indicate better shape similarity.

\subsubsection{Quality}
Quality metric evaluates the perceptual quality of the swapped face images. Following \cite{nirkin2019fsgan}, we use Structural Similarity Index (SSIM) to measure the image quality. Higher SSIM indicates better image quality.

\begin{table}
\begin{center}
\resizebox{0.9\linewidth}{!}{
\begin{tabular}{c c c c c c}
\hline
Method  & ID $\downarrow$ & Pose $\downarrow$ &  Expression $\downarrow$ & Shape $\downarrow$ &  Quality $\uparrow$ \\
\hline\hline
w/o transformer & 36.76 & 3.57 & 26.52 & 46.75 & 0.972\\
w/o multi-scale & 35.16  &  3.87  & 26.63   &  48.81 & 0.974 \\
Ours & \textbf{34.49} & \textbf{3.13} & \textbf{23.74}  & \textbf{44.3} &  \textbf{0.978}\\
\hline
\end{tabular}}
\end{center}
\vspace{-3mm}
\caption{
Quantitative ablation study of the proposed Face Transformer with versus w/o transformer and with versus w/o multi-scale feature transformation.}
\label{quantatative ablation}

\end{table}

\subsection{Quantitative Experiments}
We quantitatively evaluate and compare the proposed Face Transformer with state-of-the-art methods DeepFakes, FaceSwap and FaceShifter \cite{li2019faceshifter} over dataset Faceforensics++ \cite{rossler2019faceforensics++}.

Table \ref{quantitative} shows quantitative experimental results. It can be seen that the proposed Face Transformer outperforms the state-of-the-art in terms of \textit{ID verification}, \textit{shape}, \textit{expression} and \textit{quality}, demonstrating its clear superiority in high-fidelity and accurate face swapping. However, the proposed Face Transformer does not perform the best in pose metric (3.13 v.s. 2.51) that measures the difference between the predicted 3D pose vector of the target faces and that of the swapped faces. The lower pose accuracy is largely due to the fact that we follow \cite{li2019faceshifter} to estimate 3D pose vectors based on image intensities \cite{ruiz2018fine}. However, the proposed Face Transformer preserves better skin colours of source faces in the swapped face which actually enlarges the intensity difference between the target faces and the swapped faces and further degrades the performance under the current pose metric.

\subsection{Qualitative Experiments}

Fig. \ref{qualitative} shows qualitative experimental results. As DeepFakes and FaceSwap first synthesize inner face regions and blend them with the target face backgrounds for swapped faces, they tend to generate blending inconsistency in the synthesized faces as illustrated in the sample images in column 1, 2, 3, 6, and 7. FaceShifter \cite{li2019faceshifter} extracts identity features from source face and attribute features from target face, and adopts attention mechanism to adaptively integrate them for face swapping. However, the strong dependency on the attention mechanism often misleads the swapping generation where the identity features of swapped faces become deviated from that of source faces such as skin colours (in all sample images) and nose shapes (in sample images in column 1, 2, 4, 5). Different from FaceShifter \cite{li2019faceshifter}, the proposed Face Transformer first learns the semantic-aware correspondence between source face features and target face features and explicitly maps identity features of source faces to the corresponding regions of target faces to get the transformed features. The transformed features and the background features are then fused to directly generate the finally swapped faces. As the face features have already been mapped to the correct regions, the generation process becomes much easier and can achieve more accurate face swapping. It also removes the blending operation and hence eliminates the blending inconsistency issue effectively.

The proposed Face Transformer thus better preserve the identity of the source faces (e.g., the skin colors in sample images in column 1-5, the face features such as eyes, noses and mouth shapes in sample images in column 1, 2, 5, and makeups in sample images in column 1, 4, 5) and the attributes of the target faces (e.g., the mouth-opening degree in sample image in column 1, 4). We also evaluated the proposed Face Transformer over another two datasets FFHQ \cite{karras2019style} and VGGFace \cite{parkhi2015deep} by directly apply the trained model to the two datasets without fine-tuning. As illustrated in Fig \ref{fig1}, the proposed Face Transformer can handle face images under various conditions such as large head poses and different illumination conditions. This further demonstrates the effectiveness of our Face Transformer.

\begin{figure*}[t]
\centering
\includegraphics[width=0.85\linewidth]{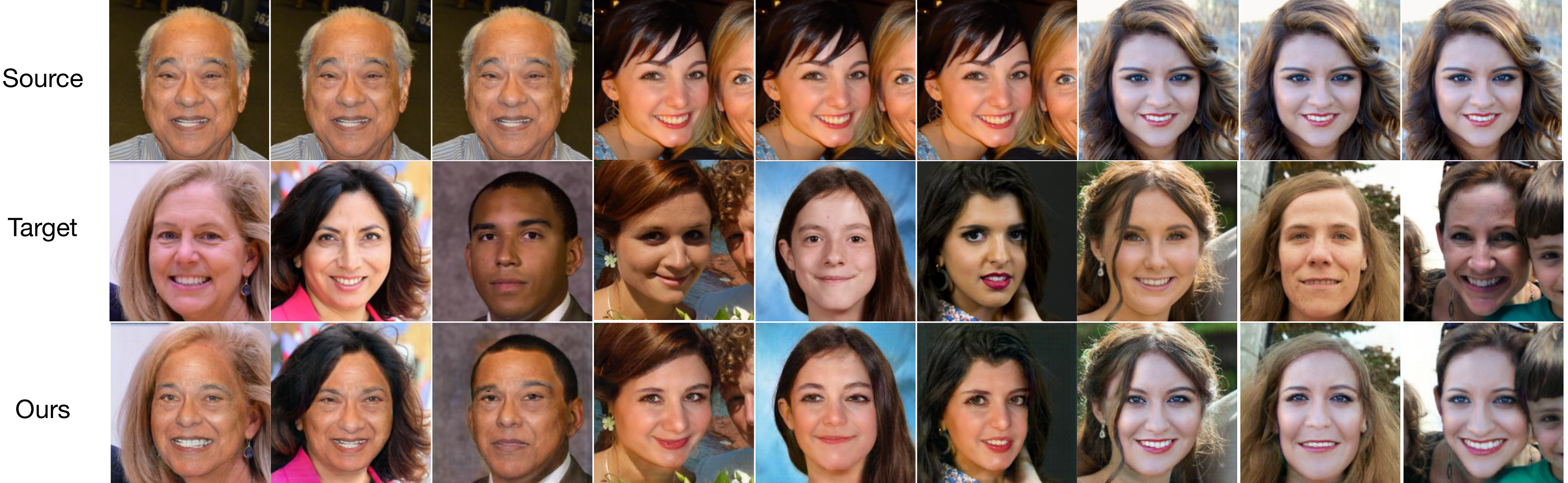}
\caption{Qualitative illustration of the proposed Face Transformer with the same source face but different target face. Every three columns form a group of images with the same source face but different target faces. It can be observed that the proposed Face Transformer produces high-fidelity and accurate face swapping.
}
\label{samesource}
\end{figure*}

\begin{figure*}[t]
\centering
\includegraphics[width=0.85\linewidth]{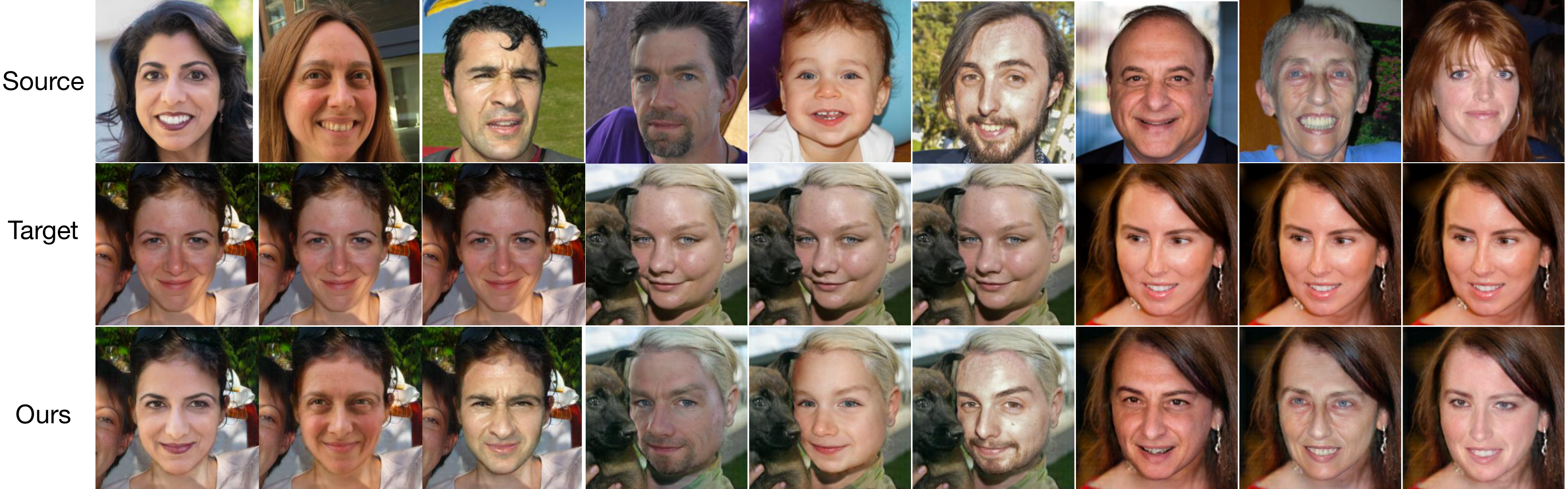}
\caption{Qualitative illustration of the proposed Face Transformer with the same target face but different source faces. Every three columns form a group of images with the same target face but different source faces. 
}
\label{sametarget}
\end{figure*}

\subsection{Ablation Study}
We perform two ablation studies quantitatively and qualitatively to demonstrate the effectiveness of our designed transformer and multi-scale feature transformation.

\subsubsection{W/o transformer} We first compare the proposed Face Transformer with a baseline model that does not employ transformer. Similar to our Face Transformer, it uses the face parsing module to separate the background and inner face, extract multi-scale features from source inner face and target background, and feeds the extracted features to our face generation module. The first row in table \ref{quantatative ablation} shows quantitative results. It can be seen that the proposed Face Transformer performs clearly better in all metrics. The better results is largely because our designed transformer accurately maps the features of source faces to their corresponding regions of target faces which improves the generation of swapped faces.

\subsubsection{W/o multi-scale}  We also compare the proposed Face Transformer with a model without using our proposed multi-scale transformation. The model uses the same transformer design but it only transforms feature $V_1$ in face generation. As shown in Table \ref{quantatative ablation}, the proposed Face Transformer performs better consistently. Without the proposed multi-scale transformation, the fidelity of swapped faces is degraded clearly.
We conclude the reason of better performance with multi-scale in two aspects. First,  multi-scale can preserve more fine facial details. Second, learning the multi-scale feature transformation can inversely help improve the learning of the semantic-aware correspondence matrix.

\subsection{Discussion}

\subsubsection{Swapping with the same source faces} Fig. \ref{samesource} shows face swapping by the proposed Face Transformer where the same source face is swapped to multiple target faces. It can be seen that the Face Transformer successfully transfers identity features of source faces to the corresponding regions of target faces while preserving the correct poses and expressions of the target faces (while target faces have very different expressions and poses).

\subsubsection{Swapping with the same target face} Fig. \ref{sametarget} shows that the proposed Face Transformer can swap different source faces to the same target faces realistically. Although the skin colours of source faces are very different, our swapped faces can preserve the skin color while maintaining its compatibility with the background by adjusting the background (e.g., the color of the neck are adjusted in the first sample). This unique feature is largely attributed to the transformer design in our Face Transformer which maps the identity features of source inner face to the corresponding regions of target faces before the generation of swapped faces. The feature transformation facilitates the learning of generation model and enables it to generate high-fidelity and realistic face swapping.

\section{Conclusion}
In this paper, we present a novel face swapping framework named Face Transformer for accurately preserving source identities and target attributes simultaneously in the swapped face images.
We introduce transformer network for the face swapping task, which learns high-quality semantic-aware correspondence between source and target faces and accurately integrates source identity information with target attributes.
In addition, our Face Transformer exploits a multi-scale transformation mechanism for preserving more fine facial details.
Extensive experiments show that the proposed Face Transformer can generate accurate and realistic face swapping results. 

\section*{Acknowledgement}
This study is funded by the Ministry of Education Singapore, under the Tier-1 scheme with a project number RG94/20.

{\small
\bibliographystyle{ieee_fullname}
\bibliography{egbib}

\begin{thebibliography}{10}\itemsep=-1pt

\bibitem{bao2017cvae}
Jianmin Bao, Dong Chen, Fang Wen, Houqiang Li, and Gang Hua.
\newblock Cvae-gan: fine-grained image generation through asymmetric training.
\newblock In {\em Proceedings of the IEEE international conference on computer
  vision}, pages 2745--2754, 2017.

\bibitem{bitouk2008face}
Dmitri Bitouk, Neeraj Kumar, Samreen Dhillon, Peter Belhumeur, and Shree~K
  Nayar.
\newblock Face swapping: automatically replacing faces in photographs.
\newblock In {\em ACM SIGGRAPH 2008 papers}, pages 1--8. 2008.

\bibitem{blanz2004exchanging}
Volker Blanz, Kristina Scherbaum, Thomas Vetter, and Hans-Peter Seidel.
\newblock Exchanging faces in images.
\newblock In {\em Computer Graphics Forum}, volume~23, pages 669--676. Wiley
  Online Library, 2004.

\bibitem{DETR}
Nicolas Carion, Francisco Massa, Gabriel Synnaeve, Nicolas Usunier, Alexander
  Kirillov, and Sergey Zagoruyko.
\newblock End-to-end object detection with transformers.
\newblock In {\em ECCV}, 2020.

\bibitem{chen2014joint}
Dong Chen, Shaoqing Ren, Yichen Wei, Xudong Cao, and Jian Sun.
\newblock Joint cascade face detection and alignment.
\newblock In {\em European conference on computer vision}, pages 109--122.
  Springer, 2014.

\bibitem{chen2020pre}
Hanting Chen, Yunhe Wang, Tianyu Guo, Chang Xu, Yiping Deng, Zhenhua Liu, Siwei
  Ma, Chunjing Xu, Chao Xu, and Wen Gao.
\newblock Pre-trained image processing transformer.
\newblock {\em arXiv preprint arXiv:2012.00364}, 2020.

\bibitem{cui2022genco}
Kaiwen Cui, Jiaxing Huang, Zhipeng Luo, Gongjie Zhang, Fangneng Zhan, and
  Shijian Lu.
\newblock Genco: generative co-training for generative adversarial networks
  with limited data.
\newblock In {\em Proceedings of the AAAI Conference on Artificial
  Intelligence}, volume~36, pages 499--507, 2022.

\bibitem{dosovitskiy2020image}
Alexey Dosovitskiy, Lucas Beyer, Alexander Kolesnikov, Dirk Weissenborn,
  Xiaohua Zhai, Thomas Unterthiner, Mostafa Dehghani, Matthias Minderer, Georg
  Heigold, Sylvain Gelly, et~al.
\newblock An image is worth 16x16 words: Transformers for image recognition at
  scale.
\newblock {\em arXiv preprint arXiv:2010.11929}, 2020.

\bibitem{esser2020taming}
Patrick Esser, Robin Rombach, and Björn Ommer.
\newblock Taming transformers for high-resolution image synthesis, 2020.

\bibitem{goodfellow2014generative}
Ian~J Goodfellow, Jean Pouget-Abadie, Mehdi Mirza, Bing Xu, David Warde-Farley,
  Sherjil Ozair, Aaron Courville, and Yoshua Bengio.
\newblock Generative adversarial networks.
\newblock {\em arXiv preprint arXiv:1406.2661}, 2014.

\bibitem{huang2022masked}
Jiaxing Huang, Kaiwen Cui, Dayan Guan, Aoran Xiao, Fangneng Zhan, Shijian Lu,
  Shengcai Liao, and Eric Xing.
\newblock Masked generative adversarial networks are data-efficient generation
  learners.
\newblock {\em Advances in Neural Information Processing Systems},
  35:2154--2167, 2022.

\bibitem{johnson2016perceptual}
Justin Johnson, Alexandre Alahi, and Li Fei-Fei.
\newblock Perceptual losses for real-time style transfer and super-resolution.
\newblock In {\em European conference on computer vision}, pages 694--711.
  Springer, 2016.

\bibitem{karras2019style}
Tero Karras, Samuli Laine, and Timo Aila.
\newblock A style-based generator architecture for generative adversarial
  networks.
\newblock In {\em Proceedings of the IEEE/CVF Conference on Computer Vision and
  Pattern Recognition}, pages 4401--4410, 2019.

\bibitem{king2009dlib}
Davis~E King.
\newblock Dlib-ml: A machine learning toolkit.
\newblock {\em The Journal of Machine Learning Research}, 10:1755--1758, 2009.

\bibitem{kingma2014adam}
Diederik~P Kingma and Jimmy Ba.
\newblock Adam: A method for stochastic optimization.
\newblock {\em arXiv preprint arXiv:1412.6980}, 2014.

\bibitem{koksal2020rf}
Ali Koksal and Shijian Lu.
\newblock Rf-gan: A light and reconfigurable network for unpaired
  image-to-image translation.
\newblock In {\em Proceedings of the Asian Conference on Computer Vision},
  2020.

\bibitem{korshunova2017fast}
Iryna Korshunova, Wenzhe Shi, Joni Dambre, and Lucas Theis.
\newblock Fast face-swap using convolutional neural networks.
\newblock In {\em Proceedings of the IEEE international conference on computer
  vision}, pages 3677--3685, 2017.

\bibitem{li2019faceshifter}
Lingzhi Li, Jianmin Bao, Hao Yang, Dong Chen, and Fang Wen.
\newblock Faceshifter: Towards high fidelity and occlusion aware face swapping.
\newblock {\em arXiv preprint arXiv:1912.13457}, 2019.

\bibitem{lin2012face}
Yuan Lin, Shengjin Wang, Qian Lin, and Feng Tang.
\newblock Face swapping under large pose variations: A 3d model based approach.
\newblock In {\em 2012 IEEE International Conference on Multimedia and Expo},
  pages 333--338. IEEE, 2012.

\bibitem{liu2021swin}
Ze Liu, Yutong Lin, Yue Cao, Han Hu, Yixuan Wei, Zheng Zhang, Stephen Lin, and
  Baining Guo.
\newblock Swin transformer: Hierarchical vision transformer using shifted
  windows.
\newblock {\em arXiv preprint arXiv:2103.14030}, 2021.

\bibitem{liu2018large}
Ziwei Liu, Ping Luo, Xiaogang Wang, and Xiaoou Tang.
\newblock Large-scale celebfaces attributes (celeba) dataset.
\newblock {\em Retrieved August}, 15(2018):11, 2018.

\bibitem{mechrez2018contextual}
Roey Mechrez, Itamar Talmi, and Lihi Zelnik-Manor.
\newblock The contextual loss for image transformation with non-aligned data.
\newblock In {\em Proceedings of the European conference on computer vision
  (ECCV)}, pages 768--783, 2018.

\bibitem{CGAN}
Mehdi Mirza and Simon Osindero.
\newblock Conditional generative adversarial nets.
\newblock {\em arXiv preprint arXiv:1411.1784}, 2014.

\bibitem{natsume2018fsnet}
Ryota Natsume, Tatsuya Yatagawa, and Shigeo Morishima.
\newblock Fsnet: An identity-aware generative model for image-based face
  swapping.
\newblock In {\em Asian Conference on Computer Vision}, pages 117--132.
  Springer, 2018.

\bibitem{natsume2018rsgan}
Ryota Natsume, Tatsuya Yatagawa, and Shigeo Morishima.
\newblock Rsgan: face swapping and editing using face and hair representation
  in latent spaces.
\newblock {\em arXiv preprint arXiv:1804.03447}, 2018.

\bibitem{nirkin2019fsgan}
Yuval Nirkin, Yosi Keller, and Tal Hassner.
\newblock Fsgan: Subject agnostic face swapping and reenactment.
\newblock In {\em Proceedings of the IEEE/CVF International Conference on
  Computer Vision}, pages 7184--7193, 2019.

\bibitem{nirkin2018face}
Yuval Nirkin, Iacopo Masi, Anh~Tran Tuan, Tal Hassner, and Gerard Medioni.
\newblock On face segmentation, face swapping, and face perception.
\newblock In {\em 2018 13th IEEE International Conference on Automatic Face \&
  Gesture Recognition (FG 2018)}, pages 98--105. IEEE, 2018.

\bibitem{park2019semantic}
Taesung Park, Ming-Yu Liu, Ting-Chun Wang, and Jun-Yan Zhu.
\newblock Semantic image synthesis with spatially-adaptive normalization.
\newblock In {\em Proceedings of the IEEE/CVF Conference on Computer Vision and
  Pattern Recognition}, pages 2337--2346, 2019.

\bibitem{parkhi2015deep}
Omkar~M Parkhi, Andrea Vedaldi, and Andrew Zisserman.
\newblock Deep face recognition.
\newblock 2015.

\bibitem{paszke2017automatic}
Adam Paszke, Sam Gross, Soumith Chintala, Gregory Chanan, Edward Yang, Zachary
  DeVito, Zeming Lin, Alban Desmaison, Luca Antiga, and Adam Lerer.
\newblock Automatic differentiation in pytorch.
\newblock 2017.

\bibitem{dcgan}
Alec Radford, Luke Metz, and Soumith Chintala.
\newblock Unsupervised representation learning with deep convolutional
  generative adversarial networks.
\newblock {\em arXiv preprint arXiv:1511.06434}, 2015.

\bibitem{rossler2019faceforensics++}
Andreas Rossler, Davide Cozzolino, Luisa Verdoliva, Christian Riess, Justus
  Thies, and Matthias Nie{\ss}ner.
\newblock Faceforensics++: Learning to detect manipulated facial images.
\newblock In {\em Proceedings of the IEEE/CVF International Conference on
  Computer Vision}, pages 1--11, 2019.

\bibitem{ruiz2018fine}
Nataniel Ruiz, Eunji Chong, and James~M Rehg.
\newblock Fine-grained head pose estimation without keypoints.
\newblock In {\em Proceedings of the IEEE conference on computer vision and
  pattern recognition workshops}, pages 2074--2083, 2018.

\bibitem{thies2016face2face}
Justus Thies, Michael Zollhofer, Marc Stamminger, Christian Theobalt, and
  Matthias Nie{\ss}ner.
\newblock Face2face: Real-time face capture and reenactment of rgb videos.
\newblock In {\em Proceedings of the IEEE conference on computer vision and
  pattern recognition}, pages 2387--2395, 2016.

\bibitem{deit}
Hugo Touvron, Matthieu Cord, Matthijs Douze, Francisco Massa, Alexandre
  Sablayrolles, and Herv\'e J\'egou.
\newblock Training data-efficient image transformers \& distillation through
  attention.
\newblock {\em arXiv preprint arXiv:2012.12877}, 2020.

\bibitem{vaswani2017attention}
Ashish Vaswani, Noam Shazeer, Niki Parmar, Jakob Uszkoreit, Llion Jones,
  Aidan~N Gomez, Lukasz Kaiser, and Illia Polosukhin.
\newblock Attention is all you need.
\newblock {\em arXiv preprint arXiv:1706.03762}, 2017.

\bibitem{wang2018cosface}
Hao Wang, Yitong Wang, Zheng Zhou, Xing Ji, Dihong Gong, Jingchao Zhou, Zhifeng
  Li, and Wei Liu.
\newblock Cosface: Large margin cosine loss for deep face recognition.
\newblock In {\em Proceedings of the IEEE conference on computer vision and
  pattern recognition}, pages 5265--5274, 2018.

\bibitem{yang2020learning}
Fuzhi Yang, Huan Yang, Jianlong Fu, Hongtao Lu, and Baining Guo.
\newblock Learning texture transformer network for image super-resolution.
\newblock In {\em Proceedings of the IEEE/CVF Conference on Computer Vision and
  Pattern Recognition}, pages 5791--5800, 2020.

\bibitem{yu2021diverse}
Yingchen Yu, Fangneng Zhan, Rongliang Wu, Jianxiong Pan, Kaiwen Cui, Shijian
  Lu, Feiying Ma, Xuansong Xie, and Chunyan Miao.
\newblock Diverse image inpainting with bidirectional and autoregressive
  transformers.
\newblock In {\em Proceedings of the 29th ACM International Conference on
  Multimedia}, pages 69--78, 2021.

\bibitem{yu2022towards}
Yingchen Yu, Fangneng Zhan, Rongliang Wu, Jiahui Zhang, Shijian Lu, Miaomiao
  Cui, Xuansong Xie, Xian-Sheng Hua, and Chunyan Miao.
\newblock Towards counterfactual image manipulation via clip.
\newblock In {\em Proceedings of the 30th ACM International Conference on
  Multimedia}, pages 3637--3645, 2022.

\bibitem{zhan2021unbalanced}
Fangneng Zhan, Yingchen Yu, Kaiwen Cui, Gongjie Zhang, Shijian Lu, Jianxiong
  Pan, Changgong Zhang, Feiying Ma, Xuansong Xie, and Chunyan Miao.
\newblock Unbalanced feature transport for exemplar-based image translation.
\newblock In {\em Proceedings of the IEEE/CVF Conference on Computer Vision and
  Pattern Recognition}, pages 15028--15038, 2021.

\bibitem{zhan2022bi}
Fangneng Zhan, Yingchen Yu, Rongliang Wu, Jiahui Zhang, Kaiwen Cui, Aoran Xiao,
  Shijian Lu, and Chunyan Miao.
\newblock Bi-level feature alignment for versatile image translation and
  manipulation.
\newblock In {\em Computer Vision--ECCV 2022: 17th European Conference, Tel
  Aviv, Israel, October 23--27, 2022, Proceedings, Part XVI}, pages 224--241.
  Springer, 2022.

\bibitem{zhan2022auto}
Fangneng Zhan, Yingchen Yu, Rongliang Wu, Jiahui Zhang, Kaiwen Cui, Changgong
  Zhang, and Shijian Lu.
\newblock Auto-regressive image synthesis with integrated quantization.
\newblock In {\em Computer Vision--ECCV 2022: 17th European Conference, Tel
  Aviv, Israel, October 23--27, 2022, Proceedings, Part XVI}, pages 110--127.
  Springer, 2022.

\bibitem{zhan2022marginal}
Fangneng Zhan, Yingchen Yu, Rongliang Wu, Jiahui Zhang, Shijian Lu, and
  Changgong Zhang.
\newblock Marginal contrastive correspondence for guided image generation.
\newblock In {\em Proceedings of the IEEE/CVF Conference on Computer Vision and
  Pattern Recognition}, pages 10663--10672, 2022.

\bibitem{zhan2019spatial}
Fangneng Zhan, Hongyuan Zhu, and Shijian Lu.
\newblock Spatial fusion gan for image synthesis.
\newblock In {\em Proceedings of the IEEE/CVF conference on computer vision and
  pattern recognition}, pages 3653--3662, 2019.

\bibitem{MetaDETR}
Gongjie Zhang, Zhipeng Luo, Kaiwen Cui, and Shijian Lu.
\newblock {Meta-DETR}: Few-shot object detection via unified image-level
  meta-learning.
\newblock {\em arXiv preprint arXiv:2103.11731}, 2021.

\bibitem{zhang2020cross}
Pan Zhang, Bo Zhang, Dong Chen, Lu Yuan, and Fang Wen.
\newblock Cross-domain correspondence learning for exemplar-based image
  translation.
\newblock In {\em Proceedings of the IEEE/CVF Conference on Computer Vision and
  Pattern Recognition}, pages 5143--5153, 2020.

\bibitem{cyclegan}
Jun-Yan Zhu, Taesung Park, Phillip Isola, and Alexei~A Efros.
\newblock Unpaired image-to-image translation using cycle-consistent
  adversarial networks.
\newblock In {\em Proceedings of the IEEE international conference on computer
  vision}, pages 2223--2232, 2017.

\bibitem{DeformableDETR}
Xizhou Zhu, Weijie Su, Lewei Lu, Bin Li, Xiaogang Wang, and Jifeng Dai.
\newblock {Deformable DETR}: Deformable transformers for end-to-end object
  detection.
\newblock In {\em ICLR}, 2021.

\end{thebibliography}
}

\end{document}